# Federated Learning Inspired Fuzzy Systems: Decentralized Rule Updating for Privacy and Scalable Decision Making

Arthur Alexander Lim [1,] Zhen Bin It[2]*, Jovan Bowen Heng[2] and Tee Hui Teo [2#]

[1]The University of Newcastle, Callaghan, 2308, Australia
[2]Singapore University of Technology and Design, Singapore
[2*]zbhienn0920@gmail.com and
[2#]tthui@sutd.edu.sg

**ABSTRACT**

Fuzzy systems are a way to allow machines, systems and frameworks to deal with uncertainty, which is not possible in binary systems that most computers use. These systems have already been deployed for certain use cases, and fuzzy systems could be further improved as proposed in this paper. Such technologies to draw inspiration from include machine learning and federated learning. Machine learning is one of the recent breakthroughs of technology and could be applied to fuzzy systems to further improve the results it produces. Federated learning is also one of the recent technologies that have huge potential, which allows machine learning training to improve by reducing privacy risk, reducing burden on networking infrastructure, and reducing latency of the latest model. Aspects from federated learning could be used to improve federated learning, such as applying the idea of updating the fuzzy rules that make up a key part of fuzzy systems, to further improve it over time. This paper discusses how these improvements would be implemented in fuzzy systems, and how it would improve fuzzy systems. It also discusses certain limitations on the potential improvements. It concludes that these proposed ideas and improvements require further investigation to see how far the improvements are, but the potential is there to improve fuzzy systems.

**Introduction**

In the modern world of high-power computing and technologies, more advanced systems and algorithms are required to make the correct decisions for more advanced and complicated scenarios. More and more details are not certain and require more considerate decision making yet also require automation to ensure very fast and reliable decision making. This means humans cannot be used to decide in all these scenarios. One of the systems that are in use today are fuzzy systems, which are designed for cases of uncertainty [1]. Fuzzy systems combine the speed of modern computing while allowing expert knowledge from humans to influence the decision making. Fuzzy systems also do not use binary, as binary only allows true or false. This system allows better decisions to be made instead of the regular binary systems, which do not factor in uncertainty and only allows absolute answers. Binary systems would only allow account strict factors and values with no values in between 2 different states. A use case for fuzzy systems includes the intrusion detection application area, where binary decision-making faces serious challenges [2]. However, like all technologies, fuzzy systems can be improved by integrating it with other new technologies. One such technology that could improve fuzzy systems under certain scenarios and use cases is federated learning. Federated learning is mostly used for machine learning models where in a wide deployment, instead of sending all the user data to one central server, it instead processes the data on the edge devices. Sending all user data creates a privacy risk [4]. This demand for machine learning with reduced privacy concerns and risk led to the development of federated learning. Once the data is processed on the edge devices, the model itself is sent back to the central server, which is smaller than the data itself and doesn't transfer the data back to another location. The aspect of distributing the models of federated learning can be adapted to fuzzy systems and can allow fuzzy systems to be improved. This paper proposes two main ideas, where the fuzzy rules themselves in the inference engine can be updated across multiple deployments of fuzzy systems, and another system where a machine learning model is paired alongside the fuzzy system to assist in its decision making. The machine learning model would then process the decisions and input data locally and then send the model back to a central server to aggregate with models from other devices. The updates to the fuzzy rules in the first idea would allow the system to improve over time and make better decisions. This paper will propose and explain further why this idea should be explored further and explains the concept in a broader sense.

**Background**

The following section will describe the background and context of what fuzzy logic is, how it works, and as well as how federated learning works and how it can be adapted to improve fuzzy logic.

**Fuzzy Logic**

Fuzzy logic is used for situations where uncertainty is present and needs to be addressed. It uses expert knowledge to create an inference engine that allows decisions to be made. For example, in intrusion detection [1], there is no clear boundary between normal and abnormal traffic that can be defined with regular systems, hence the use of fuzzy systems. Fuzzy systems take in crisp and exact values into fuzzy values, through the process of fuzzification. It categorises them into fuzzy sets, such as low, medium and high. The data is then given a value between 0 and 1 inside a fuzzy set. Fuzzy rules are then used by the system to process these values. The fuzzy rules use an IF and then statements to determine what the outcome should be given a certain fuzzy value. These are built from expert knowledge as the human knowledge with intimate knowledge can be used to build the correct and appropriate fuzzy rules for the system. The inference engine is the component of the fuzzy system that uses the established fuzzy rules to transform the inputs, which are the fuzzy values, and produce a fuzzy output. The fuzzy output is then defuzzied to convert back into a regular crisp value. The final output value is then used to influence another factor. It could be used to influence a decision-making system, or any other system.

**Federated learning**

Federated learning aims to improve the privacy and distribution aspect of machine learning models in a widespread deployment with edge devices. It does this by storing the data collected by the edge device locally, instead of transmitting the data back to a central server. The usual method of machine learning involves transmitting all the data back to a central server which then processes all the data on one location. In federated learning, the data is processed locally by the edge device itself. This improves the privacy of the data as there is no risk during transmission, and there is no one central location that contains all the data from multiple locations. Once the edge device processes the data, it updates its own machine learning model to improve the model. The machine learning models are then uploaded back to the central server, which then aggregates all the updated models to create a new updated global model. It can also be used for security purposes, such as detecting when devices are under threat from cyberattacks such as data breaches, phishing, ransomware, denial of service, and more [5]. It can also be used for other types of cyber security such as APT detection [8]. This type of learning also reduces the impact on the data transmission infrastructure, as the models are lighter and smaller than all the data generated from the data. It also allows faster response times in cyber security, as the devices have an updated model faster than the usual method of learning from a central server. Federated learning also allows a larger dataset to be used, which leads to a more accurate model, as the improved privacy means more users and organizations are willing to participate in training. This system is also more scalable as it doesn't have the same impact on the networking load as regular learning, and instead it can scale using only more edge devices, which just generates the updated models which are lighter to send.

**Methodology**

In this section of the review, the proposed improvements mentioned in the earlier section will be explained and how machine learning and federated learning could be applied to fuzzy systems in order to improve them.

**Distributing updated fuzzy rules**

As described in the earlier section of this report, one of the methods for combining aspects of federated learning and fuzzy rules include using the distributed model aspect and applying them to fuzzy rules. This allows multiple instances of deployed fuzzy systems to be updated with more accurate and desirable outcomes by using a more advanced and more accurate fuzzy rule set. Below are 2 diagrams that show which step in the fuzzy system process would be updated in this scenario.

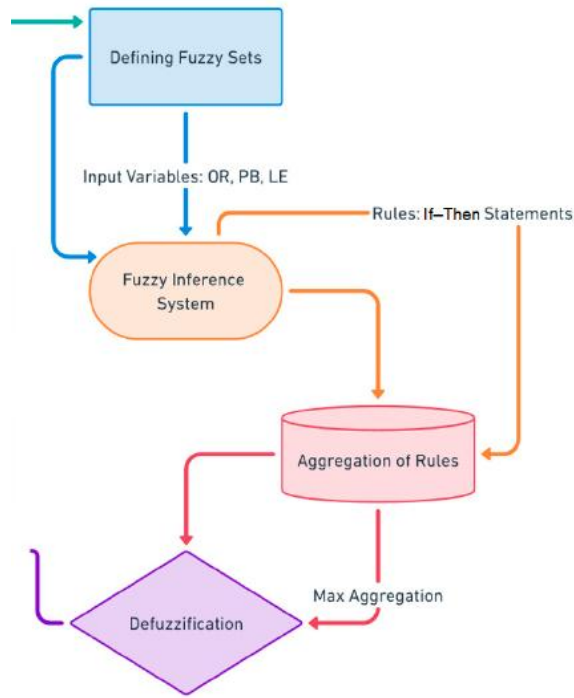

**Figure 1.** Steps of a fuzzy system [6]

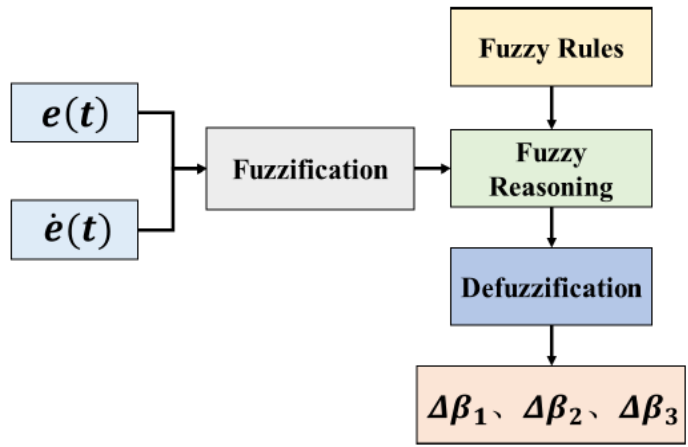

**Figure 2.** Another example of steps in a fuzzy system [6]

As these illustration shows, this updated fuzzy system will take the fuzzy inference system, specifically the part that uses the rules, which are made up of if-then statements, before aggregation of the rules and then defuzzification occurs. The if-then statements that make up the rules are derived from knowledge and experience from human experts in the field. As the general knowledge and experience of the experts increases over time, so will the need to improve and update the fuzzy rules using more if-then statements or changing them to deliver better results. And one of the methods to update the fuzzy rules is using the system from federated learning. The method in which federated learning propagates the updated models by sending the updated models back to the central server, can be applied to fuzzy systems by sending the updated fuzzy rules from the central server to the location of the deployment of the fuzzy system. And as seen in an example of a fuzzy system in the next section, there are systems that use multiple fuzzy inference systems, which will benefit even more from being able to update their fuzzy rules to create a better and more accurate system with better results.

**Adding a machine learning model alongside fuzzy system**

In this system, an additional machine learning model is used alongside the fuzzy system to improve the final result of the system. The fuzzy system would control and make the final decision making, and the machine learning model would simply assist the fuzzy system to create a more accurate and robust system. The typical use would be at the end of the There have been use cases of a system similar to this one proposed by others. In the example below, the system uses fuzzy logic to determine the risk of a ship collision in a fishing zone and creating a system to avoid such collisions [6]. The system uses fuzzy logic and fuzzy systems to create a value called a CRI, a collision risk index, a value that determines how likely a ship is to collide with another. Below is a diagram of how the fuzzy system proposed in that paper works.

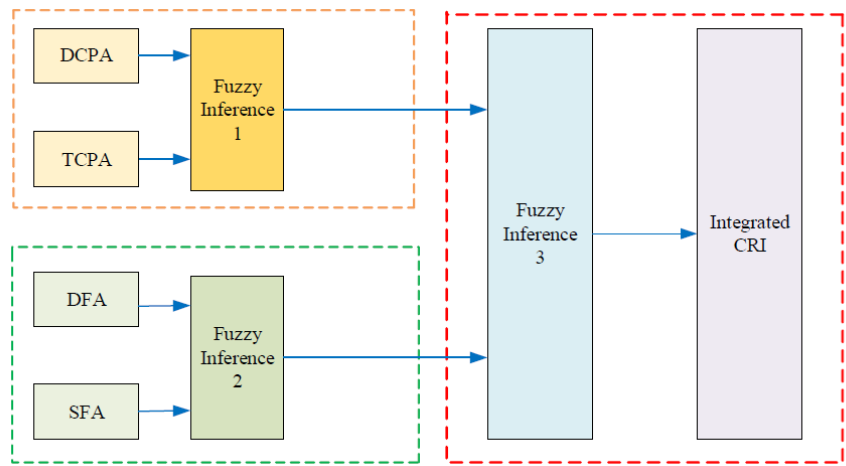

**Figure 3.** Fuzzy system deployed by the paper [7].

As it can be seen from the diagram, it uses multiple steps of fuzzy inference, which is the main component of fuzzy systems, to determine the CRI. It uses the DCPA and TCPA values determined and calculated from the distance between the two ships, and distance from fishing area (DFA) and size of fishing area (SFA). A first fuzzy inference system was used to determine a fuzzy value from the DCPA and TCPA values, and calculates how the distance between the two ships impact the risk of collision [7]. A second fuzzy inference system is used to calculate a fuzzy value from the distance from fishing are and the size of fishing area, and how those two factors would contribute to the risk of collision between ships. Then, the two fuzzy values generated from the two fuzzy inference systems are used as values for the final fuzzy inference system to generate the final collision risk index value. The final fuzzy inference system takes in the effect of the distance between the ships and the size and distance from the fishing area, to determine the final value of the collision risk index. This is how this system uses fuzzy systems to create one singular number that illustrates the collision risk [7]. This collision risk value is then used by another system, an advanced decision-making framework, to determine the optimal evasive manoeuvre so that the ships do not collide and reduce the risk of collision as much as possible. Below is the diagram of this framework.

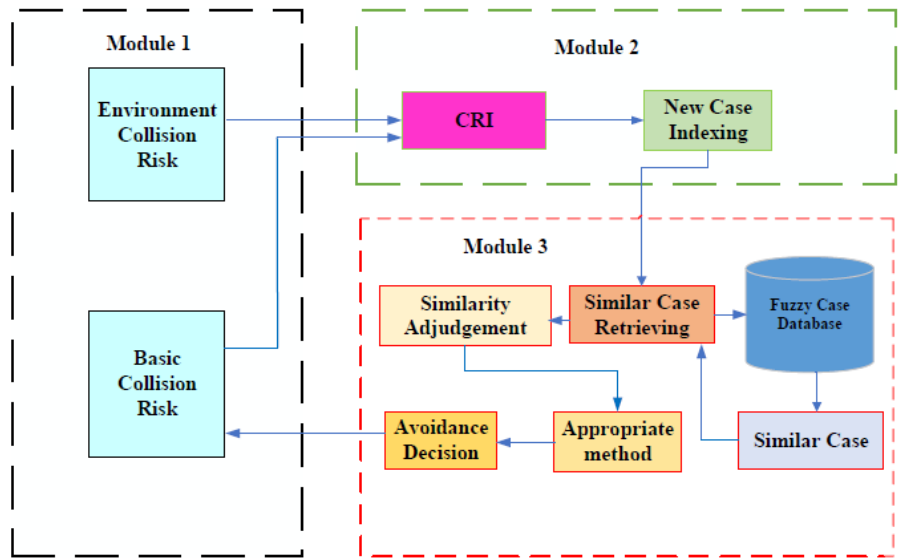

**Figure 4.** Advanced decision-making framework [7]

In this framework, module 1 calculates the basic and environment collision risk, and module 2 represents the fuzzy system shown earlier, as it generates the same collision risk index value used in module 3. This system shows that it takes the CRI value along with other factors and compares it to other scenarios in the past. It will retrieve similar cases from the past events recorded in the database and can reference a suitable avoidance strategy. If a similar case is found, it will retrieve its avoidance strategy and recommend it to the user. This aspect of the system can be improved using machine learning models to speed up the case retrieval.

It is proposed that with the introduction of machine learning models, the database can be used to train a machine learning model that can learn from all the various factors of past events along with their outcomes. This will result in a machine learning model that can understand exactly what to do in situations where there is a risk of ship collision, and what actions are required to avoid the collision, as it has been trained on a wide dataset of past events. This will speed up the decision-making process, as the machine learning model can instantly output the avoidance actions and decision. The machine learning model will in effect, replace the entire module 3, and have much less steps and require less time. A diagram of a similar system as this one is shown below.

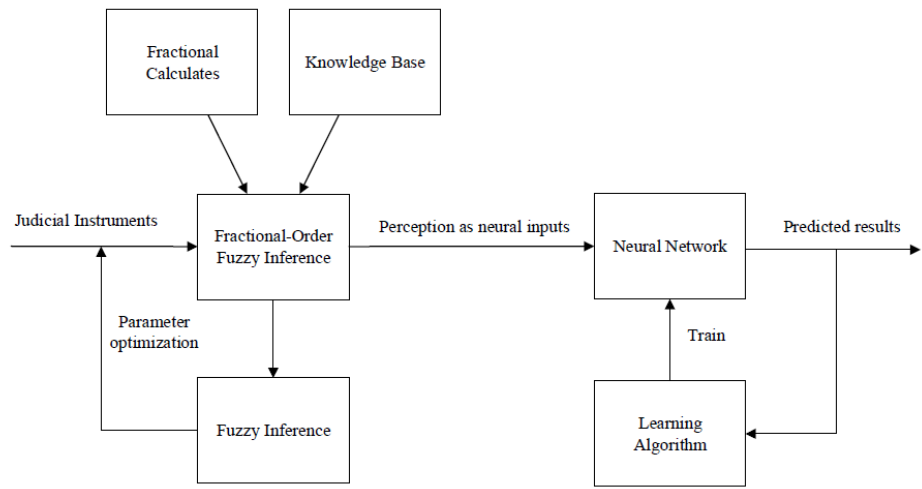

**Figure 5.** A diagram of a fuzzy system combined with neural network [9]

This system is from a different use case of fuzzy system, specifically analysing reputation infringement cases. However, it perfectly illustrates how machine learning models [10-11] can be combined with fuzzy systems to produce and predict results from the inputs. The fuzzy system functions in the first step as a normal fuzzy system would, fuzzing the inputs, using the fuzzy inference system based on the fuzzy rules derived from the knowledge base. It would then use the output of the fuzzy system as an input for the machine learning model, in this case a neural network, where it would process the input and produce an output based on the training it has done from previous results. The final result would therefore be produced based on a machine learning output after better understanding have been made from the fuzzy system.

This model is then further improved using federated learning [12-17]. The machine learning model after training with the database could then upload the updated model to a central server. From multiple instances of the machine learning model deployed in other locations, all the models are uploaded to the central server, and then aggregated into a final updated model. This updated model is then distributed back to the edge devices, which further improves the model for each instance of the machine learning model, as it has been trained on more data that isn't present on the current database.

**Discussion**

In this section, the effect and benefits of each type of proposed system are discussed, along with some limitations to highlight. As has been shown, fuzzy systems could be further improved using the federated learning models, and the updating aspect from federated learning. Updating fuzzy rules could be one of the changes that further improves fuzzy systems, as it allows the system to be more accurate and produce better results over time. Fuzzy rules are the key component in the decision-making aspect of fuzzy systems, and as human knowledge increases over time, so should the fuzzy rules that make up a fuzzy inference system. This improvement allows exponential improvement for systems with multiple fuzzy inference systems, such as the one shown earlier, which uses 3 fuzzy inference systems. Each step of the system would be drastically improved with each update to the fuzzy rules. It would also allow wide improvements to a large pool of deployed fuzzy systems, as it can update the rules to all the fuzzy systems in use.

The idea of pairing fuzzy systems with a machine learning model [18-19] and then updating the machine learning model using federated learning has huge potential. It reinforces the decision made by the fuzzy system, and it can be implemented in multiple ways. The machine learning model could be used and trained from previous use cases, or it could be used to decide the final output based on the output from the fuzzy system. These different types of applications of machine learning models allow fuzzy systems to be improved in all manners of ways, even placing the machine learning model first in order to derive a more suitable input for the fuzzy system could be an option. Alongside that, applying federated learning to said machine learning model has great improvements, such as the reduced impact on the networking infrastructure, more rapid updates to the model, and massively improved privacy risk. Without the need to upload entire databases from the location, it allows much faster improvement of the model, which in turn improves the fuzzy system. There may be potential of implementing both types of system, updating the fuzzy rules and implementing a machine learning model alongside. The limitations of this concept include the scope and practicality in certain scenarios. For more simple devices or simple scenarios, it may not be necessary to update the rules at all, nor need a machine learning model to help it. Some scenarios may not have access to a network, such as remote locations or devices that simply do not have networking capabilities. There is also a risk that the fuzzy rules may be incorrect, and therefore the fuzzy inference system degrades. This requires the party to update the fuzzy rules to implement them correctly, as unlike machine learning models training on data, human error is a possibility and may impact the system in a negative manner. Although these limitations exist, with proper care and deployment, these improvements for fuzzy systems have great potential and should be explored further.

**Conclusion**

In this paper, it has been shown how fuzzy systems could be improved by aspects of federated learning, such as the components of federated learning that allows it to update a wide deployment of systems rapidly or implementing a machine learning model itself. It's shown that these can be in several forms, such as updating the fuzzy rules itself, or using a machine learning model alongside the fuzzy system and using federated learning to improve said machine learning model. Similar concepts of applying machine learning alongside fuzzy systems have been shown and proposed by others. It was discussed how these improvements could lead to more accurate and far better results from fuzzy systems, and how these proposed improvements could work in fuzzy systems. The limitations of the improvements are also discussed, such as the scenarios where it would prove little to no benefit to the existing system, and the potential risk of human error when updating the fuzzy rules over time. However, these limitations do not make the proposed improvements not worth pursuing further, and these proposed improvements in these papers should be explored further as it could potentially lead to much better systems in certain scenarios, as well as to discover how far the improvements can take fuzzy systems to.

**Data Availability**

All data generated or analysed during this study are included in this published article.